\documentclass[10pt,twocolumn,letterpaper]{article}

\usepackage{cvpr}              

\usepackage{makecell}
\usepackage{booktabs}
\usepackage{multirow}
\usepackage{booktabs}
\usepackage{multirow}
\usepackage{colortbl}
\usepackage{makecell}
\usepackage{booktabs}
\usepackage{multirow}
\usepackage[table,dvipsnames]{xcolor} 

\definecolor{cvprblue}{rgb}{0.21,0.49,0.74}
\usepackage[pagebackref,breaklinks,colorlinks,allcolors=cvprblue]{hyperref}

\def\confName{CVPR}
\def\confYear{2026}

\title{GenAgent: Scaling Text-to-Image Generation via Agentic Multimodal Reasoning}

\author{
Kaixun Jiang\textsuperscript{1},  
Yuzheng Wang\textsuperscript{2}\textsuperscript{†},
Junjie Zhou\textsuperscript{3},  
Pandeng Li\textsuperscript{2,4}, \\
Zhihang Liu\textsuperscript{4}, 
Chen-Wei Xie\textsuperscript{2}, 
Zhaoyu Chen\textsuperscript{1}, 
Yun Zheng\textsuperscript{2}, 
Wenqiang Zhang\textsuperscript{1}  \\[0.5em] 
\textsuperscript{1}Fudan University 
\textsuperscript{2}Tongyi Lab  
\textsuperscript{3}Nanjing University 
\textsuperscript{4}University of Science and Technology of China \\ 
}

\begin{document}
\maketitle
\renewcommand{\thefootnote}{\fnsymbol{footnote}} 
\footnotetext[2]{indicates project leader.} 
\begin{abstract}
We introduce GenAgent, unifying visual understanding and generation through an agentic multimodal model. Unlike unified models that face expensive training costs and understanding-generation trade-offs, GenAgent decouples these capabilities through an agentic framework: understanding is handled by the multimodal model itself, while generation is achieved by treating image generation models as invokable tools. Crucially, unlike existing modular systems constrained by static pipelines, this design enables autonomous multi-turn interactions where the agent generates multimodal chains-of-thought encompassing reasoning, tool invocation, judgment, and reflection to iteratively refine outputs. We employ a two-stage training strategy: first, cold-start with supervised fine-tuning on high-quality tool invocation and reflection data to bootstrap agent behaviors; second, end-to-end agentic reinforcement learning combining pointwise rewards (final image quality) and pairwise rewards (reflection accuracy), with trajectory resampling for enhanced multi-turn exploration. GenAgent significantly boosts base generator(FLUX.1-dev) performance on GenEval++ (+23.6\%) and WISE (+14\%). Beyond performance gains, our framework demonstrates three key properties: 1) cross-tool generalization to generators with varying capabilities, 2) test-time scaling with consistent improvements across interaction rounds, and 3) task-adaptive reasoning that automatically adjusts to different tasks. Our code will be available at \href{https://github.com/deep-kaixun/GenAgent}{this url}.     
\end{abstract}    
\section{Introduction}
\label{sec:intro}

\begin{figure}
  \centering
  \includegraphics[width=0.46\textwidth]{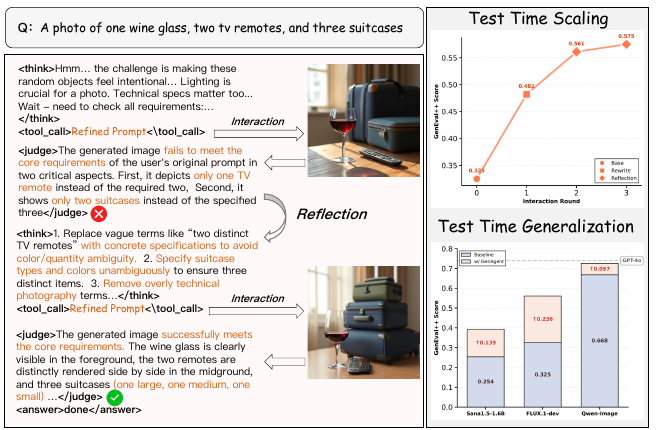}
  \vspace{-5pt}
  \caption{The left shows the illustration of GenAgent's iterative interaction with image generation tools. The right presents the performance scaling of our GenAgent across interaction rounds (top) and generalization to different image generation tools (bottom).}
  \vspace{-10pt}
  \label{fig:intro}
\end{figure}

Understanding and generation represent the two foundational pillars of multimodal intelligence~\cite{gpt4o,comanici2025gemini}. A growing body of research indicates~\cite{chen2025blip3,wu2025janus,bagel,wu2025qwen} that a tight integration of these functions can yield significant gains in complex generative tasks. Current approaches can be broadly categorized into two paradigms: integrated end-to-end models, modular and decoupled systems. The integrated approach involves either incorporating more powerful understanding modules directly into generative models~\cite{wu2025qwen,chen2025blip3} or developing unified architectures~\cite{bagel,chen2025janus,xomini,chen2025blip3o} capable of processing interleaved image-text inputs to facilitate complex reasoning chains~\cite{T2I-R1,tong2025cot,self-rewarding}. Nevertheless, such tightly-coupled designs are often constrained by performance trade-offs~\cite{niu2025wise} and require extensive computational resources for training~\cite{bagel}.

Conversely, the modular paradigm provides enhanced flexibility and extensibility by separating reasoning from generation. In this configuration, a dedicated reasoning model handles the semantic understanding, while a generative model is tasked exclusively with image rendering. RePrompt~\cite{wu2025reprompt} and PromptEnhancer~\cite{wang2025promptenhancer}, focus on single-turn prompt refinement. More recent works~\cite{wang2024genartist,iccv2025} employ pre-defined, open-loop static workflows, assigning fixed roles to a cascade of models. The performance of these systems is inherently limited by their inability to make autonomous decisions or adapt dynamically.

To address this limitation, we introduce GenAgent, an agentic framework that unifies visual understanding and generation through iterative tool interaction. GenAgent is architected around an agentic multimodal model that executes autonomous multi-turn reasoning and generation: it formulates action plans, invokes external image generators as tools, evaluates their outputs, and engages in reflective reasoning to determine whether and how to proceed. Throughout this process, the agent generates multimodal chains-of-thought encompassing reasoning, tool invocation, judgment, and reflection. This design preserves the flexibility and interpretability of the modular paradigm while enabling deep, adaptive integration of reasoning and generation capabilities. Figure~\ref{fig:intro} (left) illustrates a representative two-turn interaction.

We propose a novel two-stage training framework for GenAgent. In the first stage, we construct high-quality SFT data through hint-guided distillation. We provide the advanced model with explicit criteria and reference images to ensure it generates accurate and well-reasoned tool invocation and reflection trajectories for cold-starting multi-turn agent behaviors. In the subsequent agentic reinforcement learning (RL) stage, we introduce a hybrid reward mechanism that combines pointwise outcome rewards—which assess final image quality—with pairwise process rewards that incentivize accurate reflections across consecutive rounds. We further employ a resampling strategy on interaction rounds that oversamples diverse trajectories and uniformly resamples across different rounds to expand the exploration space. Through this paradigm, GenAgent learns autonomous multi-round tool invocation and adaptive decision-making.

We evaluate GenAgent on challenging benchmarks. Without modifying the underlying image generators, GenAgent significantly boosts the base generator (FLUX.1-dev\cite{flux2024}) performance by +23.6\% on GenEval++ and +14\% on WISE. When equipped with stronger generation tools like Qwen-Image~\cite{wu2025qwen}, our framework approaches the performance of the closed-source GPT-4o~\cite{gpt4o}. Our analysis reveals three key emergent properties: 1) \textbf{Cross-tool generalization}: GenAgent trained through interactions with FLUX.1-dev robustly transfers to other tools with varying capabilities (Figure~\ref{fig:intro}, bottom right); 2) \textbf{Test-time scaling}: performance improves consistently across interaction rounds through multi-turn refinement (Figure~\ref{fig:intro}, top right); and 3) \textbf{Task-adaptive reasoning}: distinct reasoning patterns emerge automatically for different task requirements (Section~\ref{sec:case}). We hope our agentic framework offers a promising path toward advancing unified visual understanding and generation. Our contributions are as follows:
\begin{itemize}
    \item We introduce GenAgent, an agentic multimodal model that decouples understanding and generation by treating image generators as invokable tools, enabling autonomous reasoning and interaction through multimodal chains-of-thought.
    \item We propose a two-stage training framework with cold-start SFT on curated tool-invocation and reflection trajectories and agentic RL using hybrid outcome-process rewards and resampling on interaction rounds.
    \item We demonstrate substantial gains on three challenging generation tasks, achieving competitive performance with both open-source and closed-source unified models while uncovering emergent properties including cross-tool generalization, test-time scaling, and adaptive reasoning.
\end{itemize}

\section{Related Work}
\vspace{-10pt}
\label{sec:related work}
\noindent \textbf{Understanding for generation.} 
Current approaches to enhancing understanding for controllable generation fall into two paradigms: integrated end-to-end models and modular decoupled systems. The integrated paradigm either incorporates stronger understanding modules into generative models~\cite{wu2025qwen,wan2025wan,chen2025blip3} or builds unified architectures~\cite{bagel,wu2025janus,chen2025janus,xie2024show,xomini} for complex multimodal reasoning. However, this design faces understanding-generation trade-offs and demands substantial training resources. The modular paradigm separates reasoning and generation, offering greater flexibility and lower costs. While early efforts like RePrompt~\cite{wu2025reprompt} employed simple prompt optimization, recent workflows~\cite{wang2024genartist,iccv2025} incorporate multiple models in pre-defined pipelines with fixed sequences, limiting adaptability to intermediate results. GenAgent addresses these limitations by elevating the modular paradigm from static workflow to dynamic agent. Through an agentic multimodal model-driven closed-loop mechanism, GenAgent autonomously plans, invokes tools, evaluates outputs, and iteratively refines strategies, substantially expanding the performance ceiling of modular approaches.

\noindent \textbf{Agentic multimodal reasoning.} Agentic reasoning has emerged as a powerful paradigm for complex multimodal tasks. In visual understanding, the "thinking with image" concept, pioneered by OpenAI's O3\cite{o3}, empowers models with external tools to refine reasoning—evolving from single-function tools like \cite{zheng2025deepeyes} to writing Python code for diverse image manipulations like \cite{zhang2025thyme,hong2025deepeyesv2}. In contrast, multimodal generation has predominantly relied on unified models~\cite{T2I-R1,tong2025cot,duan2025got,self-rewarding,uig}, which integrate reasoning and generation within a single architecture. However, this tightly-coupled design limits flexibility and scalability: enhancing one capability (e.g., generation quality) requires costly retraining of the entire model. Our work addresses this limitation by pioneering a decoupled agentic framework for generation. An agentic mulitimodal model serves as the reasoning agent while image generators function as callable external tools. This design enables significant advantages: generative performance can be directly improved by upgrading tool components without retraining the core agentic structure, ensuring superior flexibility and scalability.
\section{Method}

\label{sec:method}

\begin{figure*}[htb]
  \centering
  \includegraphics[width=0.95\textwidth]{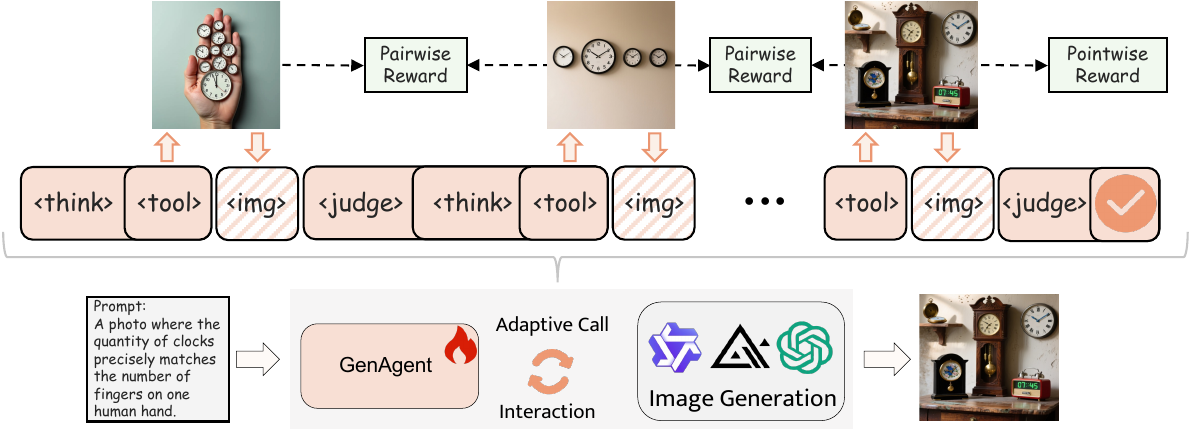}
  \caption{Overview of GenAgent. Upon receiving user input, GenAgent executes a multi-modal reasoning trace that encompasses iterative reasoning and image generation through dynamic interaction with image generation tools. During the rollout of agentic reinforcement learning, we compute pointwise rewards from the final image and pairwise rewards based on consecutive image pairs throughout trace.}
  \label{fig:main}
\end{figure*}

\subsection{Overall Pipeline}

\textbf{Overall.} Unlike existing single-step rewrite frameworks~\cite{wu2025reprompt,wang2025promptenhancer} and multi-model workflows~\cite{wang2024genartist,iccv2025}, {GenAgent} is an agentic framework for multimodal generation. As shown in Figure~\ref{fig:main}, it comprises two core components: a \textbf{multimodal model $\pi_{\theta}$} and an \textbf{interactive image generation tool}. Within each inference trajectory, GenAgent iteratively invokes the generation tool, orchestrating cycles of \textbf{thinking, generation, judgment, and reflection} until the output satisfies user requirements or reaches maximum interaction rounds $n_{\max}$. Figure~\ref{fig:main} illustrates the pipeline of our method.

\noindent\textbf{Initial Phase.} GenAgent first receives user query $q$ and reasons about the intent to produce an refined prompt $P_1$ in tool-callable format:
\begin{equation}
    T_1, P_1 = \pi_\theta(q),
\end{equation}
\noindent where $T_1$ denotes the reasoning trace. The prompt $P_1$ is passed to the image generation tool to produce $I_1 = \text{ImageGen}(P_1)$.

\noindent\textbf{Judgment and Reflection.} Then, GenAgent evaluates $I_1$ against user requirements:
\begin{equation}
  {\pi_{\theta}}(q, T_1, P_1, I_1) = 
    \begin{cases}
        (J_1, a), & \text{if } I_1 \text{ satisfies } q  \\
        (J_1, T_2, P_2), & \text{otherwise}
    \end{cases}
\end{equation}
\noindent where $J_1$ is the judgment trace and $a$ denotes termination. If the result is satisfactory, the system outputs $a$. Otherwise, $J_1$ identifies deficiencies and $T_2$, $P_2$ guide the next iteration. This process continues until satisfaction or maximum rounds $n_{\max}$ is reached. A complete multimodal reasoning trajectory (corresponding to the upper part of Figure~\ref{fig:main}) is:
\begin{equation}
    o = \{q, T_1, P_1, I_1, J_1, \ldots, T_n, P_n, I_n, J_n, a\},
\end{equation}
\noindent where $n <= n_{\max}$  denotes the actual number of interaction rounds. $o$ captures the complete multimodal chains-of-thought and self-correction capability. The final generated image $I_n$ serves as the output.

\subsection{Supervised Fine-Tuning}
\begin{table}[htb]
\centering
\caption{Diagnostic analysis of Base and SFT models on GenEval++~\cite{ye2025echo}. Error Rate represents the tool invocation failure rate, Word Diffs denotes the word count difference between the final prompt and user input, and improvement indicates the performance gain attributed to reflection.}
\resizebox{0.47\textwidth}{!}{
\begin{tabular}{@{}c|ccc@{}}
\toprule
{\color[HTML]{1F2328} Stage} & {\color[HTML]{1F2328} Error Rate(\%)} & {\color[HTML]{1F2328} Word Diffs\textless{}=5(\%)} & {\color[HTML]{1F2328} Improvement(\%)} \\ \midrule
Base                          & 13.36                                 & 49.28                                                 & 0.36                                    \\
+\textit{SFT}                           & 0.35                                 & 0.00                                                  & 6.07                                    \\ \bottomrule
\end{tabular}}
\label{tab:poor}
\end{table}

We first investigate whether multimodal models possess sufficient zero-shot capability for this task. Specifically, we evaluate Qwen2.5-VL-7B~\cite{bai2025qwen25vl} and identify three critical limitations as shown in Table~\ref{tab:poor}: 1) Unreliable tool invocation; 2) Ineffective reflection; 3) Poor refinement. These limitations indicate that direct RL would be inefficient: invocation errors prevent effective interaction with generators, weak reflection hinders learning from feedback, and poor refinement leads to limited RL exploration. Therefore, a cold-start phase is essential.

\begin{figure}[tb]
  \centering
  \includegraphics[width=0.47\textwidth]{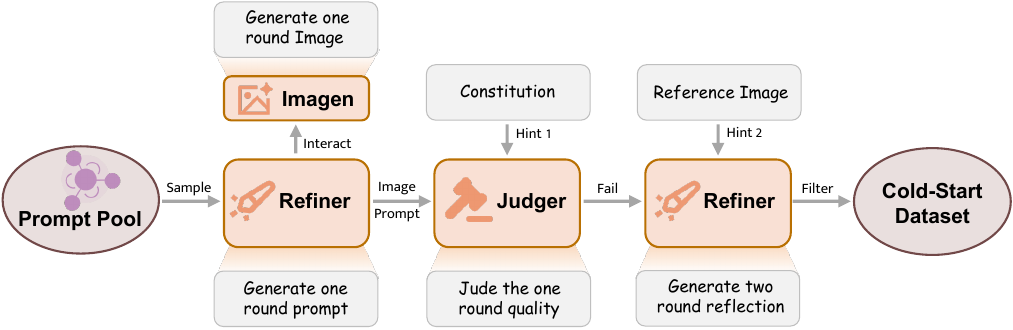}
  \caption{SFT data construction pipeline.}
  \label{fig:data_pe}
\end{figure}

\subsubsection{Data Construction Pipeline}
Figure~\ref{fig:data_pe} illustrates the complete trajectory synthesis pipeline for cold-start training data. All system prompts, data sources, and distributions are provided in the supplementary materials (Supp).

\noindent\textbf{Prompt Pool.} 
To enable the model to effectively learn tool invocation and reflection capabilities, we carefully curate sufficiently challenging training data. Specifically, we first collect open-source data, then construct synthetic data requiring complex reasoning to match the difficulty of multi-turn tool invocation (detailed pipeline in supplementary materials; Figure~\ref{fig:main} shows a synthetic case with original prompt 'a photo of five clocks'). We filter these through FLUX.1-dev~\cite{flux2024} with 3-pass generation, retaining only samples where all attempts fail.

\noindent\textbf{One-round Trajectory Synthesis.} 
The first round aims to teach the model the format for reasoning and invoking the image generation tool. We employ Qwen3-VL-235B-A22B-Thinking~\cite{qwen3vl} as the teacher model for distillation to obtain the thinking process and the refined prompt $P_1$. After filtering out format-incorrect samples, we feed all refined prompts to the FLUX.1-dev to produce the corresponding images.

\noindent\textbf{Judging Synthesis.} 
Inspired by~\cite{guan2024deliberative}, we provide explicit rules for the model to judge whether the generated image satisfies the prompt requirements and output the reasoning behind the judgment. This allows the model to implicitly learn these evaluation criteria during training, even though they are not explicitly provided during inference. We retain samples judged as successful, while unsuccessful ones proceed to the two-round reflection.

\noindent\textbf{Reflection Synthesis.}
We prompt the model to refine the one round $P_1$ based on the user's original input $q$  and previous evaluation results $J_1$. To ensure high-quality reflection, we implement two key improvements: First, we upgrade to the more capable Gemini 2.5 Pro~\cite{comanici2025gemini} as the teacher model. Second, since the seed prompt pool is derived from SFT data, all examples have corresponding reference images. We leverage these reference images as additional guidance for the second-round refinement. We filter out trajectories that explicitly reference the hint image in their prompts, then generate new images using the refined second-round prompts. Through pairwise comparison, we eliminate cases where the second round produces inferior results compared to the first. Finally, we apply pointwise evaluation to assess whether the second-round images satisfy user requirements, retaining only those samples that meet quality standards.

\noindent\textbf{Data Sampling Strategy.} 
Through the above pipeline, we construct two-round training data with tool invocation and reflection. We further filter out samples with logical inconsistencies in the trajectories. Finally, we perform balanced sampling based on data type and whether the second round terminates, ultimately obtaining 32K high-quality cold-start data samples. Notably, we intentionally avoid having all samples end with $a$ to ensure the model is not limited to single-round reflection, thereby preserving exploration space for subsequent reinforcement learning.

\subsubsection{Training Strategy}
To mitigate hallucination and leverage high-quality trajectories, we mask environment feedback tokens and compute loss only on model responses. Figure~\ref{fig:data_pe} shows that SFT successfully addresses the base model's limitations: (1) tool invocation errors decrease substantially, (2) refined prompts become longer and more detailed (see Supplementary Material), and (3) reflection quality improves markedly.

\subsection{Agentic Reinforcement Learning}

After cold-start training, GenAgent has acquired basic tool-calling capabilities. However, SFT relies on fixed round trajectories for imitation learning, making it difficult to generalize to complex scenarios requiring more rounds or to learn when to stop reflection. Therefore, we introduce reinforcement learning to train the model's dynamic decision-making abilities. We adopt GRPO~\cite{GRPO}. For each query $q$, GRPO samples a group of trajectories $\{o_1,\cdots,o_{G}\}$ from policy $\pi_\theta$ and optimizes via:
\begin{equation}
\scalebox{0.75}{$
    \begin{aligned}
    \mathcal{J}(\theta) = &
    \frac{1}{G} \sum_{i=1}^G \frac{1}{|o_i|} \sum_{t=1}^{|o_i|} \Big\{ \min \Big[
    \text{clip}(\sigma_{i,t}, 1 - \epsilon, 1 + \epsilon){A}_{i},\; \sigma_{i,t} {A}_{i}\Big] \Big\},
    \end{aligned}$}
\end{equation}
where $\sigma_{i,t} = \frac{\pi_\theta(o_{i,t} \mid q, o_{i,<t})}{\pi_{\theta_\text{old}}(o_{i,t} \mid q, o_{i,<t})}$ denotes the likelihood ratio between the updated and old policies at step $t$. $\epsilon$ is a hyperparameter. In agentic RL~\cite{zheng2025deepeyes}, we compute the loss only on the output tokens generated by $\pi_{\theta}$, while masking environment observations (i.e., the generated images $I$ in trajectory $o$). The group-normalized advantage, denoted as ${A}_{i}$, is calculated by the reward $r_{i}$:
\begin{equation}
A_{i} = \frac{r_i - \text{mean}(\{r_i\}_{i=1}^G)}{\text{std}(\{r_i\}_{i=1}^G)}.
\end{equation}
Existing RL methods~\cite{GRPO,yu2025dapo} for reasoning tasks often use rule-based outcome rewards to avoid reward hacking. However, Our task face two challenges: 1) image quality is difficult to quantify; 2) outcome rewards are too coarse-grained, failing to distinguish the quality of effective reflection trajectories. We propose a {hybrid reward mechanism} combining pointwise and pairwise rewards.

\noindent\textbf{Pointwise Reward.} We employ an advanced MLLM as a Generative Reward Model~\cite{grm} to verify whether the final generated image satisfies all specified conditions. To ensure evaluation reliability, we implement three key strategies: 1) We prompt the MLLM to first generate step-by-step reasoning before making the final judgment. 2) We provide detailed evaluation criteria along with condition-specific hints for each sample to guide accurate assessment. 3) We enforce strict rewards where the agent receives a reward only when all conditions are satisfied; otherwise, the reward is 0.

\noindent\textbf{Pairwise Reward.} To encourage effective reflection while preventing reward hacking~\cite{geirhos2020shortcut}, we introduce process rewards based on a key insight: genuine reflection should yield consistent quality improvement across the entire trajectory. We grant rewards only when the MLLM judges that all subsequent images are superior to their predecessors:
\begin{equation}
r_{\text{pair}} = \begin{cases}
0.3 & \text{if } M_{\text{judge}}(I_{k}, I_{k+1}) = I_{k+1}, \forall k \in [1, n-1], \\
0 & \text{otherwise},
\end{cases}
\end{equation}
where $M_{\text{judge}}(I_{k}, I_{k+1})$ returns the better image between consecutive pairs. To avoid positional bias, the positions are randomly shuffled. The final reward is: 
\begin{equation}
r= r_{\text{point}} + r_{\text{format}} +  \lambda \cdot r_{\text{pair}},
\end{equation}
where $r_{\text{point}} \in \{0,0.7\}, r_{\text{format}} \in \{0,-0.2\},$ $\lambda =0.5$ if  $r_{\text{point}}=0$ else $\lambda =1$.

\noindent \textbf{Resampling on Interaction Rounds.} Motivated by~\cite{shang2025rstar2}, considering that trajectories with different numbers of interaction turns correspond to distinct reasoning traces (\textit{e.g.}, single-turn interactions primarily focus on rewriting and judging, while multi-turn interactions involve reflection), we adopt a resampling strategy based on interaction turns. Specifically, during the rollout phase, we oversample $G'$ trajectories ($G' > G$), and then uniformly resample according to the number of interaction turns. This approach enables the model to observe more diverse trajectories and expands the exploration space.

\section{Experiments}
\label{sec:exp}

\subsection{Implenmention Details}
We initialize our approach using Qwen2.5-VL-7B~\cite{bai2025qwen25vl} as the base model and perform supervised fine-tuning via the LLaMAFactory~\cite{zheng2024llamafactory}, employing a batch size of 256 and a learning rate of $1\times10^{-5}$. The model undergoes optimization across 3 epochs utilizing the AdamW optimizer. In the subsequent reinforcement learning phase, we implement an enhanced GRPO with a batch size of 240. For each prompt, we generate $G'=12$ rollouts, which are then downsampled to $G=8$ samples. The KL divergence coefficient is set to 0.0~\cite{yu2025dapo}, the maximum response length is capped at 16,384 tokens, and a learning rate of $1\times10^{-6}$ is adopted. We leverage the 8-step distilled version of FLUX.1-dev~\cite{flux-distilled} as our interactive image generation tool, maximum number of interactions $n_{max}=3$; all subsequent references to FLUX.1-dev correspond to this specific version. Through efficient and robust server deployment, we substantially improve rollout efficiency. We employ Qwen3-VL-30B-A3B~\cite{qwen3vl} as the generative reward model. The complete training framework is implemented using verl~\cite{verl}.

\subsection{ Benchmarks}
We select three challenging benchmarks to evaluate our method across multiple dimensions: 
\textbf{(1) Instruction-following:} GenEval++~\cite{ye2025echo}, a more accurate and challenging benchmark for evaluating instruction fidelity in image generation, featuring richer semantics and more diverse compositions than the original GenEval~\cite{ghosh2023GenEval}. 
\textbf{(2) Knowledge-grounded reasoning:} WISE~\cite{niu2025wise}, which comprehensively assesses complex semantic understanding and world-knowledge integration in text-to-image generation. 
\textbf{(3) Creative generation:} Imagine~\cite{ye2025echo}, designed to evaluate surreal and imaginative image generation capabilities by testing the model's ability to augment common objects with fantastical elements while preserving their core identity—often requiring outputs that contradict factual reality.

\begin{table*}[htb]
\centering
\caption{Performance Comparison on GenEval++. \textbf{bold} indicates the best result.}
\vspace{-5pt}
\resizebox{0.9\textwidth}{!}{
\begin{tabular}{@{}c|c|cccccccc}
\toprule
{\color[HTML]{1F2328} Type}      & {\color[HTML]{1F2328} Method}                 & {\color[HTML]{1F2328} Color}  & {\color[HTML]{1F2328} Count}  & {\color[HTML]{1F2328} Color/Count} & {\color[HTML]{1F2328} Color/Pos} & {\color[HTML]{1F2328} Pos/Count} & {\color[HTML]{1F2328} Pos/Size} & {\color[HTML]{1F2328} Multi-Count} & {\color[HTML]{1F2328} Overall} \\ \midrule
                                 & FLUX.1-dev                                   & 0.400                         & 0.600                         & 0.250                              & 0.250                            & 0.075                            & 0.400                           & 0.300                              & 0.325                          \\
\multirow{-2}{*}{Diffusion}      & Qwen-Image                                   & \textbf{0.875}                         & 0.725                         & \textbf{0.725}                              & 0.600                            & 0.475                            & \textbf{0.725}                           & 0.550                              & 0.668                          \\ \midrule
                                 & GPT4o                                        & \textcolor{gray}{ 0.900}  & \textcolor{gray}{0.675}  & \textcolor{gray}{ 0.725}       & \textcolor{gray}{ 0.625}     & \textcolor{gray}{ 0.600}     & \textcolor{gray}{ 0.800}    & \textcolor{gray}{ 0.850}       & \textcolor{gray}{ 0.739}   \\
                                 & Janus Pro 7B                                 & 0.450                         & 0.300                         & 0.125                              & 0.300                            & 0.075                            & 0.350                           & 0.125                              & 0.246                          \\
                                 & T2I-R1                                       & 0.675                         & 0.325                         & 0.200                              & 0.350                            & 0.075                            & 0.250                           & 0.300                              & 0.311                          \\
\multirow{-4}{*}{Unified}        & Bagel                                        & 0.325                         & 0.600                         & 0.250                              & 0.325                            & 0.250                            & 0.475                           & 0.375                              & 0.371                          \\ \midrule
                                  & PromptEnhancer                               & 0.500                         & 0.625                         & 0.225                              & 0.375                            & 0.125                            & 0.450                           & 0.375                              & 0.382                          \\
                                 & \multicolumn{1}{l|}{ReflectionFLow}          & 0.400                         & 0.625                         & 0.275                              & 0.275                            & 0.200                            & 0.425                           & 0.325                              & 0.361                          \\ 
\multirow{-3}{*}{Decoupled}   & \cellcolor[HTML]{EFEFEF}Ours  \textit{w/o} tool      & \cellcolor[HTML]{EFEFEF}0.650 & \cellcolor[HTML]{EFEFEF}0.575 & \cellcolor[HTML]{EFEFEF}0.350      & \cellcolor[HTML]{EFEFEF}0.450    & \cellcolor[HTML]{EFEFEF}0.350    & \cellcolor[HTML]{EFEFEF}0.600   & \cellcolor[HTML]{EFEFEF}0.400      & \cellcolor[HTML]{EFEFEF}0.482  \\ \midrule 
                                 & {Base}                                         & 0.300                         & 0.600                         & 0.325                              & 0.300                            & 0.325                            & 0.300                           & 0.350                              & 0.357                          \\
                                 & \cellcolor[HTML]{EFEFEF}\textit{+SFT}        & \cellcolor[HTML]{EFEFEF}0.500 & \cellcolor[HTML]{EFEFEF}0.550 & \cellcolor[HTML]{EFEFEF}0.550      & \cellcolor[HTML]{EFEFEF}0.525    & \cellcolor[HTML]{EFEFEF}0.250    & \cellcolor[HTML]{EFEFEF}0.600   & \cellcolor[HTML]{EFEFEF}0.575      & \cellcolor[HTML]{EFEFEF}0.507  \\
                                 & \cellcolor[HTML]{EFEFEF}\textit{+RL}         & \cellcolor[HTML]{EFEFEF}{0.750} & \cellcolor[HTML]{EFEFEF}0.675 & \cellcolor[HTML]{EFEFEF}0.375     & \cellcolor[HTML]{EFEFEF}0.525    & \cellcolor[HTML]{EFEFEF}0.450    & \cellcolor[HTML]{EFEFEF}0.625   & \cellcolor[HTML]{EFEFEF}0.525      & \cellcolor[HTML]{EFEFEF}0.561  \\
\multirow{-4}{*}{\makecell{GenAgent\\(Ours)}} & \cellcolor[HTML]{EFEFEF}\textit{w/} {Qwen-Image} & \cellcolor[HTML]{EFEFEF}0.775 & \cellcolor[HTML]{EFEFEF}\textbf{0.775} & \cellcolor[HTML]{EFEFEF}0.650      & \cellcolor[HTML]{EFEFEF}\textbf{0.800}    & \cellcolor[HTML]{EFEFEF}\textbf{0.600}    & \cellcolor[HTML]{EFEFEF}\textbf{0.725}   & \cellcolor[HTML]{EFEFEF}\textbf{0.750}      & \cellcolor[HTML]{EFEFEF}\textbf{0.725}  \\ \bottomrule
\end{tabular}}
\end{table*}

\begin{table*}[ht]
\centering
\caption{Performance Comparison on WISE. \textbf{bold} indicates the best result.}
\vspace{-5pt}
\resizebox{0.8\textwidth}{!}{
\begin{tabular}{c|c|ccccccc}
\toprule
{\color[HTML]{1F2328} Type} & {\color[HTML]{1F2328} Method} & {\color[HTML]{1F2328} Cultural} & {\color[HTML]{1F2328} Time} & {\color[HTML]{1F2328} Space} & {\color[HTML]{1F2328} Biology} & {\color[HTML]{1F2328} Physics} & {\color[HTML]{1F2328} Chemistry} & {\color[HTML]{1F2328} Overall} \\ \midrule
\multirow{2}{*}{Diffusion}                                & {FLUX.1-dev }                 & 0.56                            & 0.57                        & 0.67                         & 0.45                           & 0.54                           & 0.41                             & 0.55                           \\
     & Qwen-Image                   & 0.62                            & 0.63                        & 0.77                         & 0.57                           & 0.75                           & 0.40                             & 0.62                           \\ \midrule
                                 & GPT4o                        & \textcolor{gray}{0.81}                            &\textcolor{gray}{0.71}                        & \textcolor{gray}{0.89}                         & \textcolor{gray}{0.83}                           & \textcolor{gray}{0.79}                           & \textcolor{gray}{0.74}                             & \textcolor{gray}{0.80}                           \\
                                 & Janus Pro 7B                 & 0.30                            & 0.37                        & 0.49                         & 0.36                           & 0.42                           & 0.26                             & 0.35                           \\
                                 & T2I-R1                       & 0.56                            & 0.55                        & 0.63                         & 0.54                           & 0.55                           & 0.30                             & 0.54                           \\
     
 & Bagel                        & 0.44                            & 0.55                        & 0.68                         & 0.44                           & 0.60                           & 0.39                             & 0.52                           \\
 \multirow{-5}{*}{Unified}  
  & Bagel \textit{w/} Self-CoT                        & 0.76                           & \textbf{0.69}                        & 0.75                         & 0.65                           & \textbf{0.75}                           & \textbf{0.58}                             & 0.70                           \\ \midrule
                                 & PromptEnhancer               & 0.54                            & 0.60                        & 0.69                         & 0.46                           & 0.62                           & 0.39                             & 0.56                           \\
\multirow{-2}{*}{Decoupled}  & \cellcolor[HTML]{EFEFEF}Ours  \textit{w/o} tool               & \cellcolor[HTML]{EFEFEF}0.75                            & \cellcolor[HTML]{EFEFEF}0.66                        & \cellcolor[HTML]{EFEFEF}0.72                         & \cellcolor[HTML]{EFEFEF}0.55                           & \cellcolor[HTML]{EFEFEF}0.60                           & \cellcolor[HTML]{EFEFEF}0.49                             & \cellcolor[HTML]{EFEFEF}0.67                           \\  \midrule
                                 & Base                      & 0.70                            & 0.63                        & 0.68                         & 0.53                           & 0.61                           & 0.41                             & 0.63                           \\
                                 & \cellcolor[HTML]{EFEFEF}\textit{+SFT}               & \cellcolor[HTML]{EFEFEF}0.68                            & \cellcolor[HTML]{EFEFEF}0.61                        & \cellcolor[HTML]{EFEFEF}0.68                         & \cellcolor[HTML]{EFEFEF}0.60                           & \cellcolor[HTML]{EFEFEF}0.65                           & \cellcolor[HTML]{EFEFEF}0.47                             & \cellcolor[HTML]{EFEFEF}0.64                           \\
                                 & \cellcolor[HTML]{EFEFEF}  \textit{+RL}                   & \cellcolor[HTML]{EFEFEF}0.75                            & \cellcolor[HTML]{EFEFEF}\textbf{0.69}                        & \cellcolor[HTML]{EFEFEF}0.72                         & \cellcolor[HTML]{EFEFEF}0.61                           & \cellcolor[HTML]{EFEFEF}0.65                           & \cellcolor[HTML]{EFEFEF}0.50                             & \cellcolor[HTML]{EFEFEF}0.69                           \\
\multirow{-4}{*}{\makecell{GenAgent\\(Ours)}}  & \cellcolor[HTML]{EFEFEF} \textit{w/} {Qwen-Image}         & \cellcolor[HTML]{EFEFEF}\textbf{0.78}                            & \cellcolor[HTML]{EFEFEF}0.67                        & \cellcolor[HTML]{EFEFEF}\textbf{0.78}                         & \cellcolor[HTML]{EFEFEF}\textbf{0.72}                           & \cellcolor[HTML]{EFEFEF}{0.71}                           & \cellcolor[HTML]{EFEFEF}{0.55}                             & \cellcolor[HTML]{EFEFEF}\textbf{0.72}                           \\ \bottomrule
\end{tabular}}
\end{table*}

\begin{table*}[htb]
\centering
\caption{Performance Comparison on Imagine. \textbf{bold} indicates the best result.}
\vspace{-5pt}
\resizebox{0.8\textwidth}{!}{
\begin{tabular}{@{}c|c|ccccc}
\toprule
{\color[HTML]{1F2328} Type}      & {\color[HTML]{1F2328} Method}                 & {\color[HTML]{1F2328} Attribute Shift} & {\color[HTML]{1F2328} Spatiotemporal} & {\color[HTML]{1F2328} Hybridization} & {\color[HTML]{1F2328} Multi-Object} & {\color[HTML]{1F2328} Overall} \\ \midrule
                                 & FLUX.1-dev                                  & 5.298                                  & 6.350                                 & 7.053                                & 5.973                               & 6.072                          \\
\multirow{-2}{*}{Diffusion}      & Qwen-Image                                   & 6.771                                  & 7.193                                 & 8.130                                & 7.500                               & 7.329                          \\ \midrule
                                 & GPT4o                                        & \textcolor{gray}{ 8.540}           & \textcolor{gray}{ 9.180}          & \textcolor{gray}{ 8.570}         & \textcolor{gray}{ 7.980}        & \textcolor{gray}{ 8.560}   \\
                                 & Janus Pro 7B                                 & 5.300                                  & 7.280                                 & 6.730                                & 6.040                               & 6.220                          \\
                                 & T2I-R1                                       & 5.850                                  & 7.700                                 & 7.360                                & 6.680                               & 6.780                          \\
\multirow{-4}{*}{Unified}        & Bagel                                        & 5.370                                  & 6.930                                 & 6.500                                & 6.410                               & 6.200                          \\ \midrule
                                 & PromptEnhancer                               & 5.489                                  & 6.213                                 & 7.327                                & 6.493                               & 6.281                          \\
\multirow{-2}{*}{Decoupled}   & \cellcolor[HTML]{EFEFEF}Ours  \textit{w/o} tool      & \cellcolor[HTML]{EFEFEF}5.829          & \cellcolor[HTML]{EFEFEF}6.827         & \cellcolor[HTML]{EFEFEF}7.130        & \cellcolor[HTML]{EFEFEF}6.137       & \cellcolor[HTML]{EFEFEF}6.408  \\ \midrule
                                 & Base                                         & 5.467                                  & 6.590                                 & 7.287                                & 6.083                               & 6.258                          \\
                                 & \cellcolor[HTML]{EFEFEF}\textit{+SFT}        & \cellcolor[HTML]{EFEFEF}6.282          & \cellcolor[HTML]{EFEFEF}7.033         & \cellcolor[HTML]{EFEFEF}7.480        & \cellcolor[HTML]{EFEFEF}6.527       & \cellcolor[HTML]{EFEFEF}6.770  \\
                                 & \cellcolor[HTML]{EFEFEF}\textit{+RL}         & \cellcolor[HTML]{EFEFEF}6.293          & \cellcolor[HTML]{EFEFEF}7.067         & \cellcolor[HTML]{EFEFEF}7.543        & \cellcolor[HTML]{EFEFEF}6.663       & \cellcolor[HTML]{EFEFEF}6.825  \\
\multirow{-4}{*}{\makecell{GenAgent\\(Ours)}} & \cellcolor[HTML]{EFEFEF}\textit{w/} {Qwen-Image} & \cellcolor[HTML]{EFEFEF}\textbf{7.613}          & \cellcolor[HTML]{EFEFEF}\textbf{7.547}         & \cellcolor[HTML]{EFEFEF}\textbf{8.343}        & \cellcolor[HTML]{EFEFEF}\textbf{7.763}       & \cellcolor[HTML]{EFEFEF}\textbf{7.794}  \\ \bottomrule
\end{tabular}}
\end{table*}

\subsection{Performance Comparison}
We selected representative baselines spanning different paradigms: 1) Diffusion models: Qwen-Image~\cite{wu2025qwen} and FLUX-1.dev~\cite{flux2024}; 2) Unified architectures: the autoregressive Janus-Pro-7B~\cite{chen2025janus}, reasoning-focused T2I-R1~\cite{T2I-R1}, hybrid-architecture Bagel~\cite{bagel}, and closed-source GPT-4o~\cite{gpt4o}; 3) Decoupled approaches: PromptEnhancer (7B version)~\cite{wang2025promptenhancer} and ReflectionFlow~\cite{iccv2025}. These methods and GenAgent use FLUX-1.dev as the default image generation tool. For fair comparison and efficiency, we report results with $n_{\max}=2$ for all multi-turn baselines and our method. We report our GenAgent across multiple variants: without tool invocation, degenerating to a single-pass rewrite model (\textit{w/o} tool), with SFT (\textit{+SFT}), with RL (\textit{+RL}), and with a stronger tool (\textit{w/} {Qwen-Image}).

\noindent \textbf{Diffusion Models.} GenAgent substantially enhances diffusion model performance across all benchmarks. When FLUX.1-dev is equipped with our GenAgent framework, it achieves substantial improvements: +0.14 on WISE (0.69 vs. 0.55), +0.236 on GenEval++ (0.561 vs. 0.325), and +0.753 on ImagineBench (6.825 vs. 6.072). Notably, despite RL training with FLUX, GenAgent generalizes effectively to the stronger Qwen-Image generator, achieving 0.72, 0.725, and 7.803—new state-of-the-art results among open-source methods. This demonstrates weak-to-strong scalability at the tool level, analyzed in detail in Section~\ref{sec:tool}.

\noindent \textbf{Unified Architectures.} On knowledge-grounded reasoning (WISE), GenAgent \textit{w/} FLUX.1-dev achieves 0.69, matching the specialized Bagel \textit{w/} Self-CoT (0.70), while GenAgent \textit{w/} Qwen-Image reaches 0.72, surpassing all open-source unified models and approaching GPT-4o (0.80). For instruction-following (GenEval++), GenAgent with Qwen-Image (0.725) nearly matches GPT-4o (0.739) and significantly outperforms other open-source alternatives. On creative generation (Imagine), GenAgent with Qwen-Image (7.794) substantially narrows the gap with GPT-4o (8.560) while establishing a clear lead over the second-best T2I-R1 (6.780). These results demonstrate that our agentic framework achieves superior generalization and scalability compared to unified architectures, while maintaining competitive performance on complex generative tasks.

\noindent \textbf{Decoupled Methods.} GenAgent substantially outperforms existing decoupled approaches. Single-rewrite PromptEnhancer achieves only marginal gains over vanilla FLUX: 0.56 on WISE (+0.01), 0.382 on GenEval++ (+0.057), and 6.281 on Imagine (+0.209). Workflow-based ReflectionFlow reaches 0.361 on GenEval++, significantly below GenAgent's 0.561 (+0.2). Notably, even without the agentic framework, using only one rewriting (Ours \textit{w/o} tool), our method outperforms PromptEnhancer across all benchmarks. These results indicate that existing decoupled methods underutilize multimodal model reasoning capabilities. In contrast, GenAgent's end-to-end agentic loop with iterative judgment and reflection unlocks significant potential.

\subsection{Ablation Study}

\begin{table}[ht]
\caption{Ablation study on training stages. Subscripts represent the gains attributed to reflection.}
\centering
\vspace{-5pt}
\resizebox{0.45\textwidth}{!}{
\begin{tabular}{@{}c|ccc@{}}
\toprule
Stage     & WISE                    & GenEval++                 & Imagine                   \\ \midrule
Base       & 0.63                    & 0.357                     & 6.258                     \\
SFT        & 0.64                    & 0.507                     & 6.770                     \\
RL \textit{w/o} $r_{\text{pair}}$ & $0.68_{\color{orange}+0.01}$          & $0.543_{\color{orange}+0.068}$          & $6.814_{\color{orange}+0.168}$          \\
RL(Ours) & \textbf{$\mathbf{0.69}_{{\color{orange}+0.02}}$} & \textbf{$\mathbf{0.561}_{{\color{orange}+0.079}}$} & \textbf{$\mathbf{6.825}_{{\color{orange}+0.417}}$} \\ \bottomrule
\end{tabular}}
\label{tab:aba_stage}
\end{table}

\begin{figure}[t]
  \centering
  \includegraphics[width=0.4\textwidth]{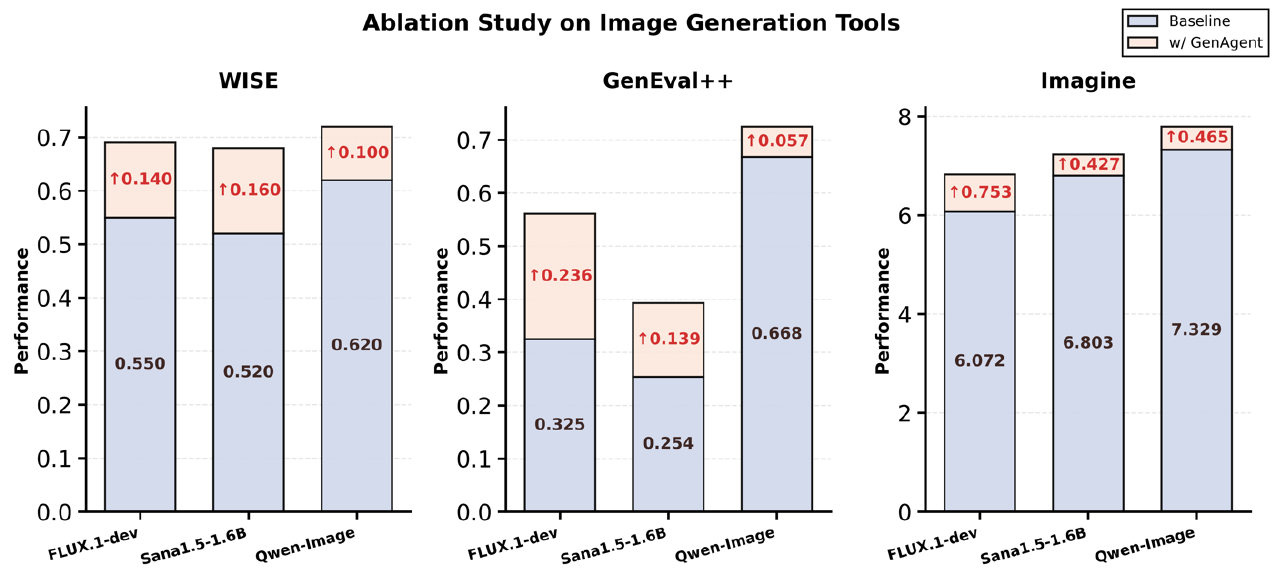}
  \vspace{-5pt}
  \caption{Ablation study on image generation tools.}
  \vspace{-10pt}
  \label{fig:aba_tools}
\end{figure}

\textbf{Training Stage.} We conduct ablation studies to assess the contribution of each training stage (Table~\ref{tab:aba_stage}). The initial Supervised Fine-Tuning (SFT) stage improves instruction-following ({GenEval++}) and creativity metrics ({Imagine}) but yields limited gains on the knowledge-intensive {WISE} benchmark. This is because SFT primarily focuses on learning "how to respond" rather than exploring to activate the model's intrinsic reasoning capabilities. The subsequent RL phase ({RL w/o $r_{\text{pair}}$}) improves performance across all metrics, demonstrating that RL-based exploration can effectively elicit the model's latent reasoning capabilities. Most critically, the full model incorporating our reflection reward, {RL(Ours)}, achieves the strongest performance overall. Compared to its counterpart without the pairwise reward, our method delivers substantially greater reflection-induced improvements, which translate directly into superior downstream performance. This validates the effectiveness of our $r_{\text{pair}}$ design in amplifying the value of reflection within the agentic framework.

\begin{figure}[t]
  \centering
  \includegraphics[width=0.4\textwidth]{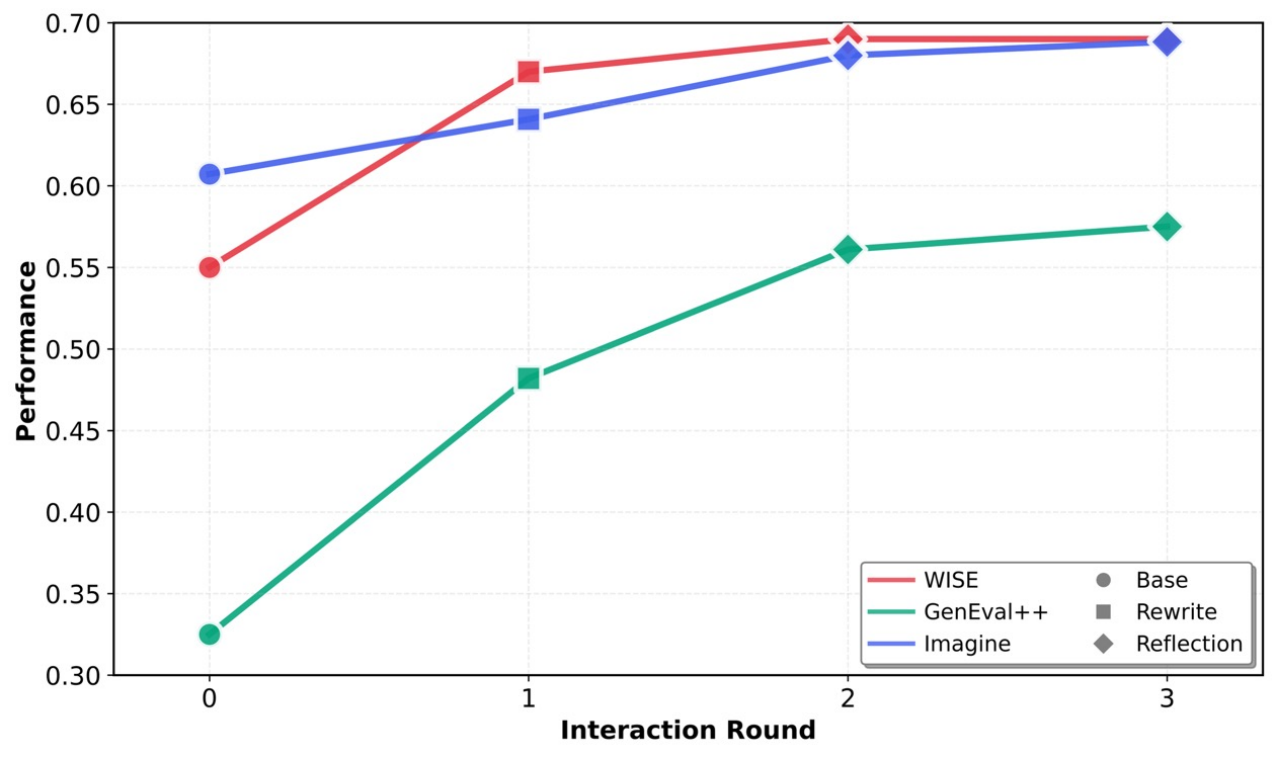}
  \vspace{-5pt}
  \caption{Ablation on interaction rounds, with normalized scores.}
  \label{fig:aba_inter}
\end{figure}

\noindent \textbf{Image Generation Tools. \label{sec:tool}} As illustrated in Figure~\ref{fig:aba_tools}, although our GenAgent uses FLUX.1-dev as the image generation tool during reinforcement learning, it demonstrates exceptional cross-tool generalization and scalability. When replacing the tool with the small Sana1.5-1.6B model~\cite{xieSANA15Efficient2025}, GenAgent still achieves significant improvements across all three benchmarks, demonstrating strong backward compatibility. More encouragingly, when employing the more powerful Qwen-Image, GenAgent further enhances its overall performance. This phenomenon indicates that our method can scale with the capability improvements of underlying image generation tools, fully unleashing the potential of advanced generative models.

\begin{figure}[t]
  \centering
  \includegraphics[width=0.47\textwidth]{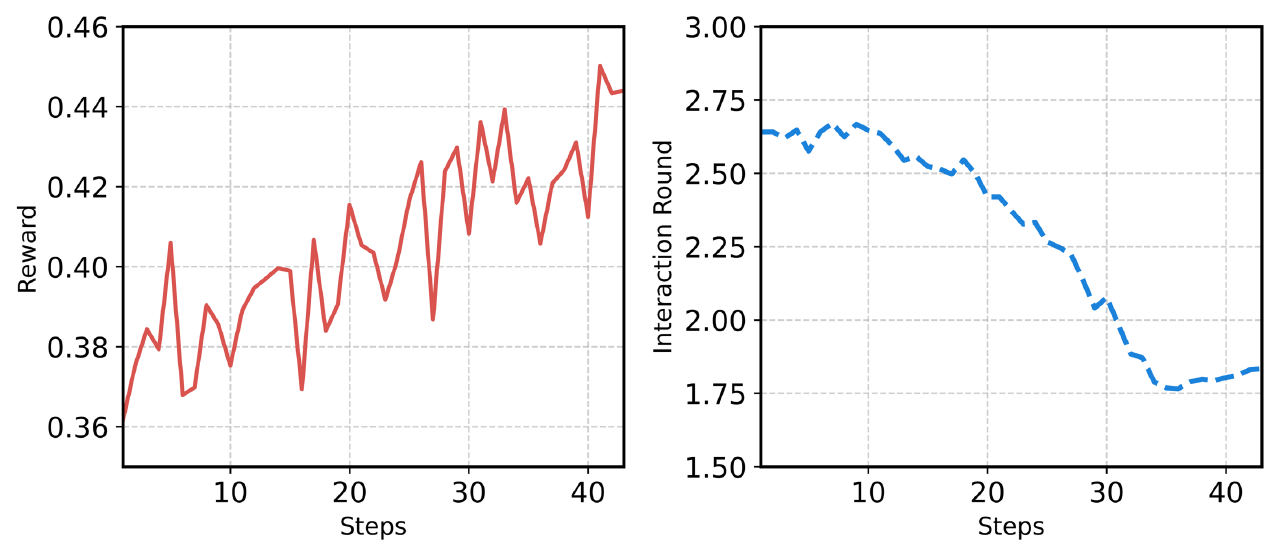}
  \vspace{-5pt}
  \caption{Training dynamics on reward and interaction rounds.}
  \label{fig:training}
\end{figure}

\begin{figure*}[ht]
  \centering
  \includegraphics[width=0.99\textwidth]{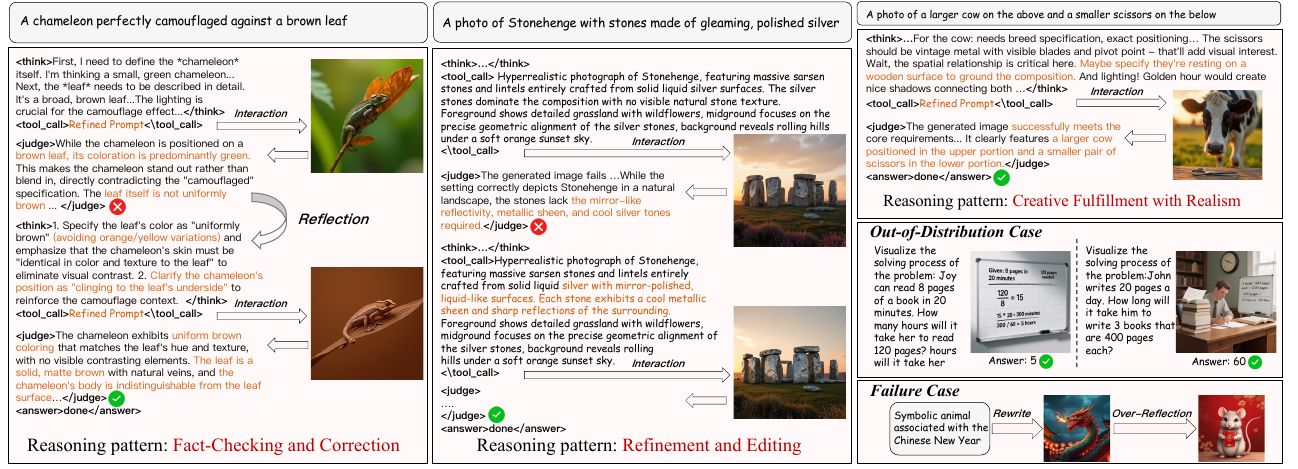}
  \caption{Case Studies of GenAgent. The figure illustrates three distinct reasoning patterns of GenAgent, its out-of-distribution generalization on a visual math task, and a representative failure case.}
  \label{fig:vis}
\end{figure*}

\noindent \textbf{Interaction Rounds.}
Figure~\ref{fig:aba_inter} illustrates the impact of different interaction rounds with the image generation tool on performance. The first interaction demonstrates GenAgent's rewriting capability, with all benchmarks exhibiting significant performance improvements, indicating that initial prompt rewriting effectively comprehends and refines users' original prompts. The second interaction round showcases the effectiveness of the first reflection, with all three tasks showing varying degrees of performance gains, confirming that the reflection mechanism successfully optimizes prompts through introspective analysis. By the third interaction round (i.e., the second reflection), while the performance gains are more modest compared to the first round, further improvements remain achievable. Our analysis reveals that one contributing factor is the inherent capability ceiling of the underlying image generation model, which possesses intrinsic limitations in understanding complex semantics and fine-grained attributes that cannot be overcome through further prompt optimization. The experimental results indicate that two interaction rounds represent the optimal balance between performance and computational efficiency.

\noindent \textbf{Training Dynamic.} We further analyze model dynamics during RL training by tracking reward progression and the number of interaction rounds with the image generation tool, as illustrated in Figure~\ref{fig:training}. In the initial phase, the model exhibits a steady decrease in interactions with the image generation tool, suggesting that it first learns to leverage the rewriting capability rather than immediately engaging in multiple reflection rounds. This behavior corrects the over-reflection tendency inherited from the supervised fine-tuning phase. Subsequently, the model begins to increase interaction rounds to obtain pairwise rewards. As training progresses, both interaction rounds and rewards rise synchronously, demonstrating that the model has successfully learned when and how to perform effective reflection.

\subsection{Case Study \label{sec:case}}
\textbf{Key Reasoning Patterns.} We highlight three key reasoning patterns in GenAgent's iterative process:

\noindent\textbf{1) Fact-Checking and Correction:} In the chameleon example (Figure~\ref{fig:vis}, left), the agent detects a logical conflict---a green chameleon on brown leaves contradicts ``perfectly camouflaged''---and revises the instruction to align color and texture, demonstrating self-correction capability.

\noindent\textbf{2) Refinement and Editing:} For near-correct outputs like Stonehenge (Figure~\ref{fig:vis}, mid), GenAgent performs targeted edits rather than full regeneration, adding specific descriptors (e.g., ``mirror-polished'') to efficiently capture missing details like stone gloss.

\noindent\textbf{3) Creative Fulfillment with Realism:} The cow-and-scissors case (Figure~\ref{fig:vis}, right-top) shows GenAgent balancing creativity with plausibility by constructing a convincing scene through composition, lighting, and perspective.

\noindent \textbf{Out-of-Distribution Cases.} Beyond these patterns, we test GenAgent on out-of-distribution tasks. Inspired by~\cite{tong2025thinking}, we design a mathematical visualization task using GSM8K~\cite{gsm8k} problems (Figure~\ref{fig:vis}, right-mid): GenAgent first reasons through the problem, then converts solution steps into detailed visual prompts, demonstrating strong potential for multimodal deep research.

\noindent \textbf{Failure Case.} We also identify a failure mode: {Over-Reflection}, where ambiguous prompts cause the agent to cycle through valid interpretations. 

Further qualitative analyses, visual comparisons between different methods, and additional complete cases are provided in Supplementary Materials.

\section{Conlcusion}
In this work, we explore how to construct agentic multimodal models that unify visual understanding and generation through active tool invocation. We introduce GenAgent, which decouples these capabilities—understanding is handled by the multimodal model itself, while generation leverages external models as tools. We conduct a two-stage training pipeline: SFT on high-quality tool invocation and reflection data, followed by reinforcement learning with pointwise and pairwise rewards to enable iterative refinement. Our analysis reveals task-adaptive reasoning and cross-tool generalization, with reinforcement learning enabling complex multi-turn interactions. Extensive experiments demonstrate that our framework significantly enhances the generation performance of traditional diffusion models, achieving results comparable to powerful unified models, while maintaining greater flexibility and lower training costs. We hope our work offers an alternative perspective for advancing unified understanding and generation in multimodal models.
\label{sec:conlusion}

{
    \small
    \bibliographystyle{ieeenat_fullname}
    \bibliography{main}
}

\end{document}